\documentclass[a4paper,conference]{IEEEtran}

\usepackage{color}
\usepackage{cite}
\usepackage{graphicx}
\usepackage{subfigure}
\usepackage {amsmath}
\usepackage {booktabs}
\usepackage {array}
\usepackage{tabularx}
\usepackage{capt-of}
\usepackage{float}
\usepackage{caption}
\usepackage{subfig}
\usepackage{setspace}
\usepackage{bm}
\usepackage{marvosym}

\usepackage{multirow}
\usepackage{amsfonts,amssymb}
\usepackage{lineno} 

\ifCLASSINFOpdf

\else

\fi

\IEEEoverridecommandlockouts

\begin{document}

\title{Multimodal Tree Decoder for Table of Contents Extraction in Document Images}

\author{\IEEEauthorblockN{Pengfei Hu\textsuperscript{1}, Zhenrong Zhang\textsuperscript{1}, Jianshu Zhang\textsuperscript{2}, Jun Du\textsuperscript{1,*}\thanks{* Corresponding Author},  Jiajia Wu\textsuperscript{2}}
\IEEEauthorblockA{\textsuperscript{1}National Engineering Research Center of Speech and Language Information Processing\\
University of Science and Technology of China, Hefei, Anhui, P. R. China\\
\textsuperscript{2}iFLYTEK Research\\
Email: hudeyouxiang@mail.ustc.edu.cn, zzr666@mail.ustc.edu.cn\\ jszhang6@iflytek.com, jundu@ustc.edu.cn,  jjwu@iflytek.com}
}

\maketitle
\begin{abstract}
Table of contents (ToC) extraction aims to extract headings of different levels in documents to better understand the outline of the contents, which can be widely used for document understanding and information retrieval. Existing works often use hand-crafted features and predefined rule-based functions to detect headings and resolve the hierarchical relationship between headings. Both the benchmark and research based on deep learning are still limited. Accordingly, in this paper, we first introduce a standard dataset, HierDoc, including image samples from 650 documents of scientific papers with their content labels. Then we propose a novel end-to-end model by using the multimodal tree decoder (MTD) for ToC as a benchmark for HierDoc. The MTD model is mainly composed of three parts, namely encoder, classifier, and decoder. The encoder fuses the multimodality features of vision, text, and layout information for each entity of the document. Then the classifier recognizes and selects the heading entities. Next, to parse the hierarchical relationship between the heading entities, a tree-structured decoder is designed. To evaluate the performance, both the metric of tree-edit-distance similarity (TEDS) and F1-Measure are adopted. Finally, our MTD approach achieves an average TEDS of 87.2\% and an average F1-Measure of 88.1\% on the test set of HierDoc. The code and dataset will be released at: https://github.com/Pengfei-Hu/MTD.


\end{abstract}

\IEEEpeerreviewmaketitle

\section{INTRODUCTION}
\begin{figure}
\centerline{\includegraphics[width=1\linewidth]{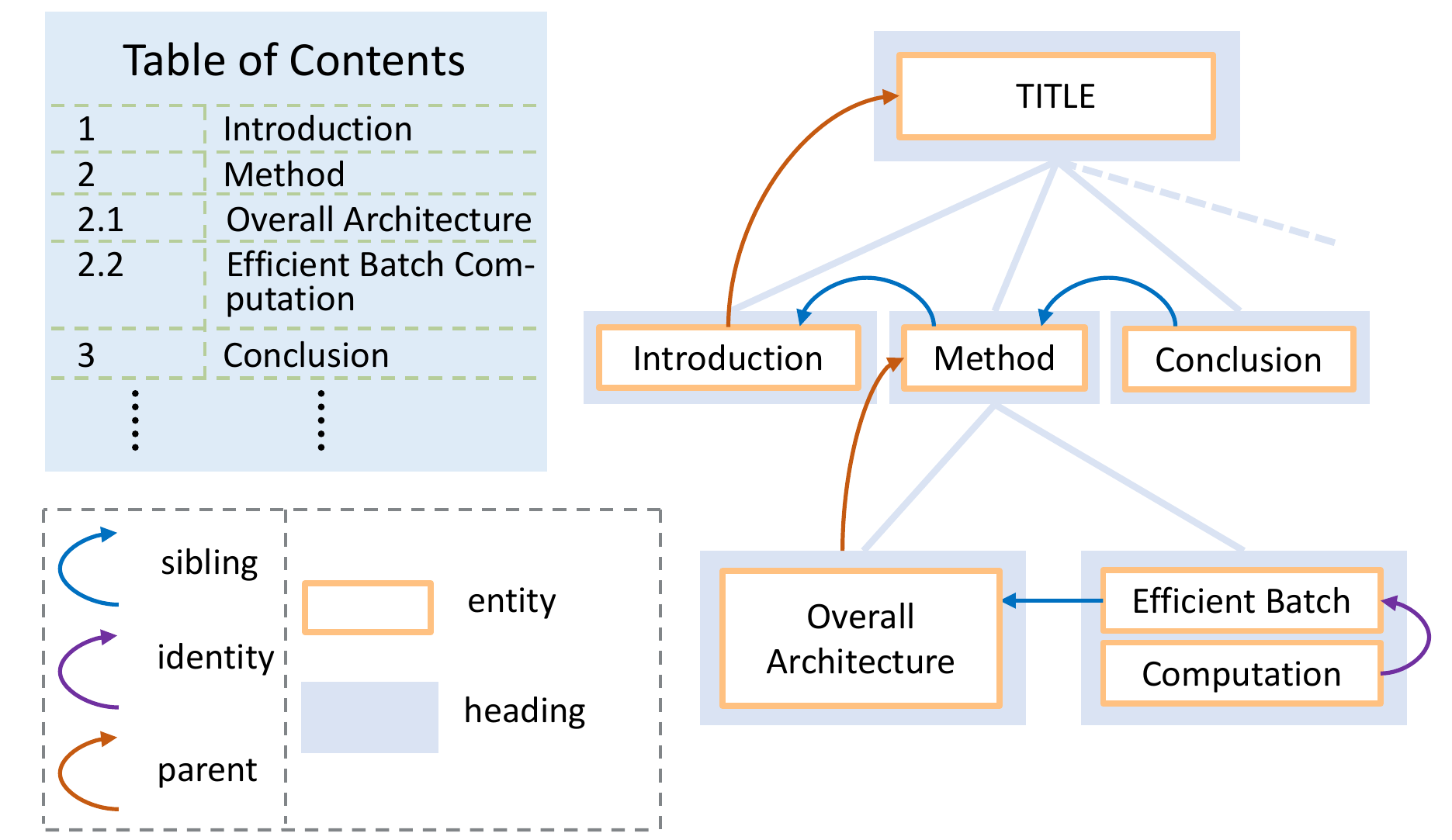}}
\caption{The ToC can be represented as a tree structure, and the output of MTD can be easily converted to ToC. The heading ``Efficient Batch Computation" is split into two entities in the document due to the limited length of the text line.}
\label{figure:formal}
\end{figure}

A huge amount of documents have been accumulated with digitization and OCR engines. However, most of them contain text with limited structural information. For example, the table of contents (ToC), which plays an important role in document understanding and information retrieval, is often missing. The task of ToC extraction is to restore the structure of the document and to recognize the hierarchy of sections. As shown in Fig. \ref{figure:formal}, the output of ToC extraction is a tree of headings with different levels.

As for the recent studies, we can roughly divide them into two categories. The first one assumes the presence of ToC pages\cite{ 2020Tocpages } \cite{ 2017Tocpages}. They first detect ToC pages, then analyze them for ToC entries. However, there are quite a few documents without ToC pages\cite{noToC}. Therefore, others extract ToC from the whole document\cite{2018content} \cite{2007content}. They usually utilize hand-crafted features and strong indicators to detect headings, which will be hierarchically ordered according to predefined rule-based functions. Besides, some research explores hybrid approaches\cite{2015hybrid}. They first consider whether the document contains ToC pages, then apply one of the methods above.

In general, existing approaches depend greatly on strong indicators and predefined rule-based functions. They can perform well on application-dependent and domain-specific datasets. However, a large amount of task-specific knowledge and human-designed rules are needed, which does not extend to other types of documents. Recently, deep learning based methods have achieved great success in many fields related to documents. For example, \cite{layoutlm} proposes the LayoutLM method for document image understanding, which is inspired by BERT\cite{devlin2018bert}. This method uses image features and position to pre-train the model and performs well on downstream tasks. Lately, ViBERTgrid\cite{vibertgrid} is proposed for key information extraction from documents. These demonstrate the powerful ability of deep learning based methods to deal with document-related problems.

It is worth noting that the existing datasets are not suitable for deep learning based methods. The ISRI dataset \cite{isri} and the Medical Article Records Groundtruth (MARG) dataset \cite{marg} only contain bi-level images and similarly simple layouts, predominantly of journal articles. Some datasets \cite{2015icdar} \cite{2017icdar} \cite{2009icdar} are available in several ICDAR challenges, which contain complex layouts of newspapers, books, and technical articles. However, their annotations lack the heading category. Annotations with the heading category are provided by PubLayNet \cite{publaynet}. However, there is no information about the heading depth. \cite{data:fin} collects 71 French documents and 72 English documents in PDF format from the financial domain. Its structure extraction ground truth is still not aligned with the text lines in the document.

To overcome the lack of data for the research of ToC extraction, we collect a dataset, namely \textbf{Hier}archical academic \textbf{Doc}ument (\textit{HierDoc}). It contains 650 academic English documents in various fields from ArXiv\footnote{https://arxiv.org/}. With LaTeX source code, we generate its ToC ground truth using regular expressions. We also provide annotations for each text line. In this paper, the text line is denoted as the entity. The ToC is aligned with entities for training. For this dataset, 350 and 300 documents are used for training and testing individually. More details of the dataset are described in Section \ref{sec:dataset}.

With \textit{HierDoc}, it is possible to take advantage of deep neural networks for the ToC extraction task. We further propose \textbf{M}ultimodal \textbf{T}ree \textbf{D}ecoder (MTD), an end-to-end model which detects heading entities and produces the ToC by parsing the hierarchy of them. To the best of our knowledge, this is the first deep learning based method for ToC extraction. Different from previous works, MTD offers the following advantages, (1) it can process documents in images or PDF format, and (2) no human-designed rules are used. It indicates that the MTD can generalize well across documents from different domains and handle different forms of user input. We adopt the text line as the basic input element, denoted as entity. MTD mainly has three components: encoder, classifier and decoder. Firstly, the encoder as a feature extractor embeds vision, plain text, and layout of the entity into a feature vector. Then the classifier detects heading entities and feeds them to the decoder. Finally, the decoder predicts the relationships between the heading entities one by one. More specifically, the attention mechanism built into the decoder locates the reference entity of a heading entity, and the relationship between the two is predicted afterward. The output of the MTD can be transferred to the ToC simply, as shown in Fig. \ref{figure:formal}. We utilize both the Tree-Edit-Distance-based Similarity (TEDS)\cite{teds} metric and the F1-Measure to evaluate the performance of our model. MTD achieves an average TEDS of 87.2\% and an average F1-Measure of 88.1\% on the test set of \textit{HierDoc}. The ablation studies prove the effectiveness of each module of MTD.

The main contributions of this paper are as follows:

\begin{itemize}
\item We introduce a standard dataset  \textit{HierDoc} for ToC extraction, which contains 650 documents of scientific papers from various fields.
\item We propose a novel end-to-end model, Multimodal Tree Decoder (MTD) for ToC extraction. We demonstrate that fusing visual, textual, and layout features boosts model performance.
\item By predicting the relationships between entities, the hierarchy of headings can be sequentially parsed by the tree decoder in MTD.
\item We achieve the results with an average TEDS of 87.2\% and an average F1-Measure of 88.1\% on the test set of \textit{HierDoc}, which provides a competitive baseline for subsequent research.
\end{itemize}

\section{DATASET\label{sec:dataset}}
As discussed in Section I, existing datasets \cite{2015icdar, 2017icdar,2009icdar,publaynet, data:fin} are not suitable for deep learning based methods. To overcome the lack of training data, we release a dataset \textit{HierDoc} which contains 350 and 300 document images for training and testing, respectively. As illustrated in Fig. \ref{figure:dataset}, \textit{HierDoc} provides two kinds of annotations, document-level annotations and entity-level annotations. The document-level annotation serves as the target of ToC extraction, which is used in the test phase. Entity-level annotations are created for training. For each entity in the document, we generate a quadruple, $(content, position, heading, id)$. $content$ is the textual content of the entity. $position$ is defined by $( x_0, y_0, x_1, y_1)$, where $( x_0, y_0 )$ corresponds to the position of the upper left in the bounding box, and $( x_1, y_1 )$ represents the position of the lower right. $heading$ is a boolean variable indicating whether the entity is a heading (or part of a heading, see Fig. \ref{figure:dataset} (b)). $id$ represents the hierarchical position in the ToC tree. For example, for entity $t$ with $content$ ``How do CR fluxes vary with Galactic..." in the Fig. \ref{figure:dataset} (a), $id$ is ``2.2", which indicates that $t$ is the second child of its parent with $content$ ``Science case" and $id$ ``2". And $t$ has one child with the $content$ ``Mechanical structure" and $id$ ``2.2.1".

\subsection{Document Collection} 
We download 650 English scientific documents in PDF format and corresponding LaTeX source codes from arXiv, covering 8 fields including physics, mathematics, computer science, quantitative biology, etc. Each document is free to distribute, remix, and adapt under the Creative Commons Attribution 4.0 user license\footnote{https://creativecommons.org/licenses}.

\subsection{Label Generation}
\begin{figure}
	\centering  
	\subfigbottomskip=2pt 
	\subfigcapskip=0pt 
	\subfigure[ducoment-level annotations]{
		\includegraphics[width=0.9\linewidth]{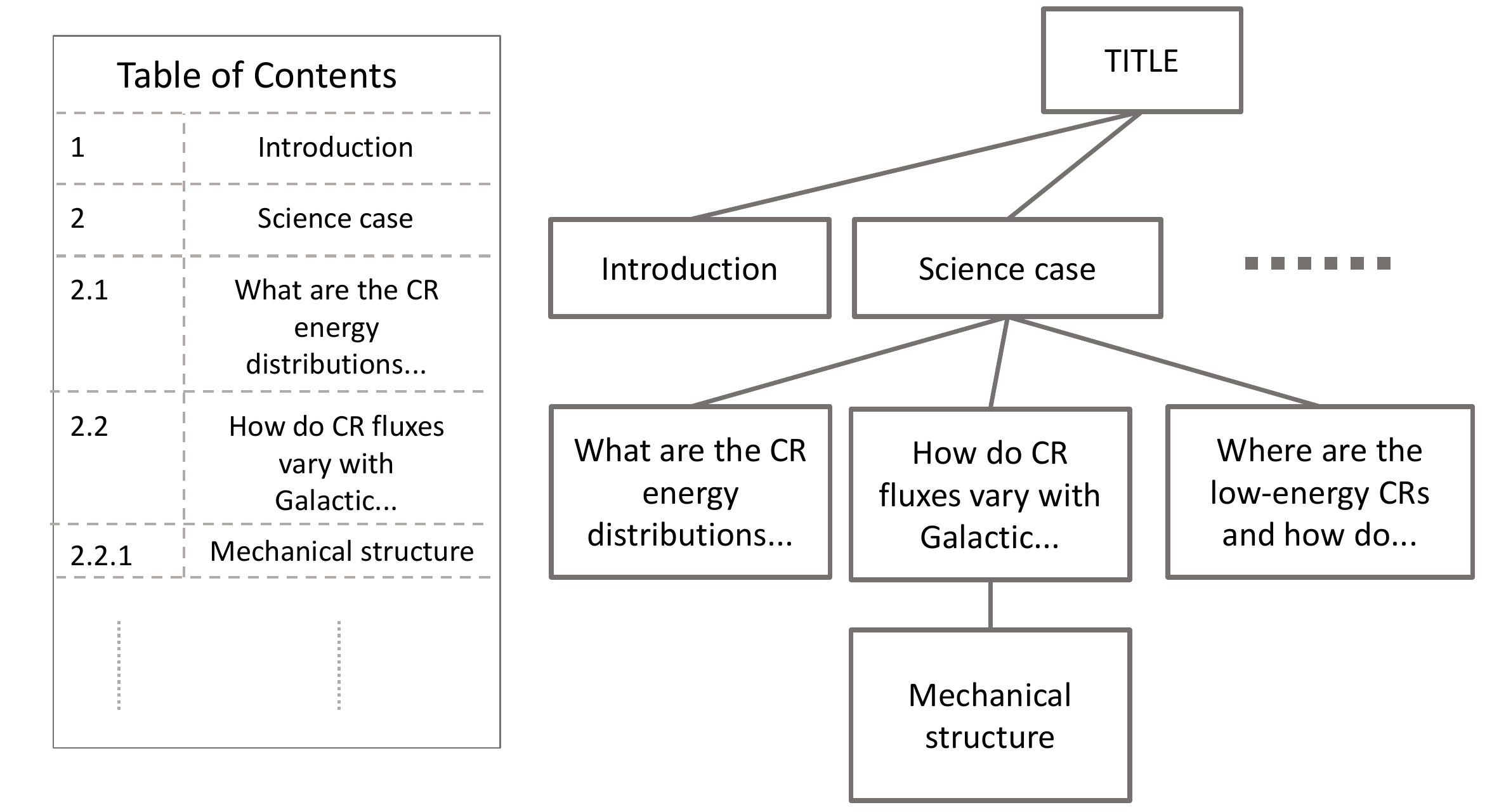}}
	\subfigure[entity-level annotations]{
		\includegraphics[width=0.9\linewidth]{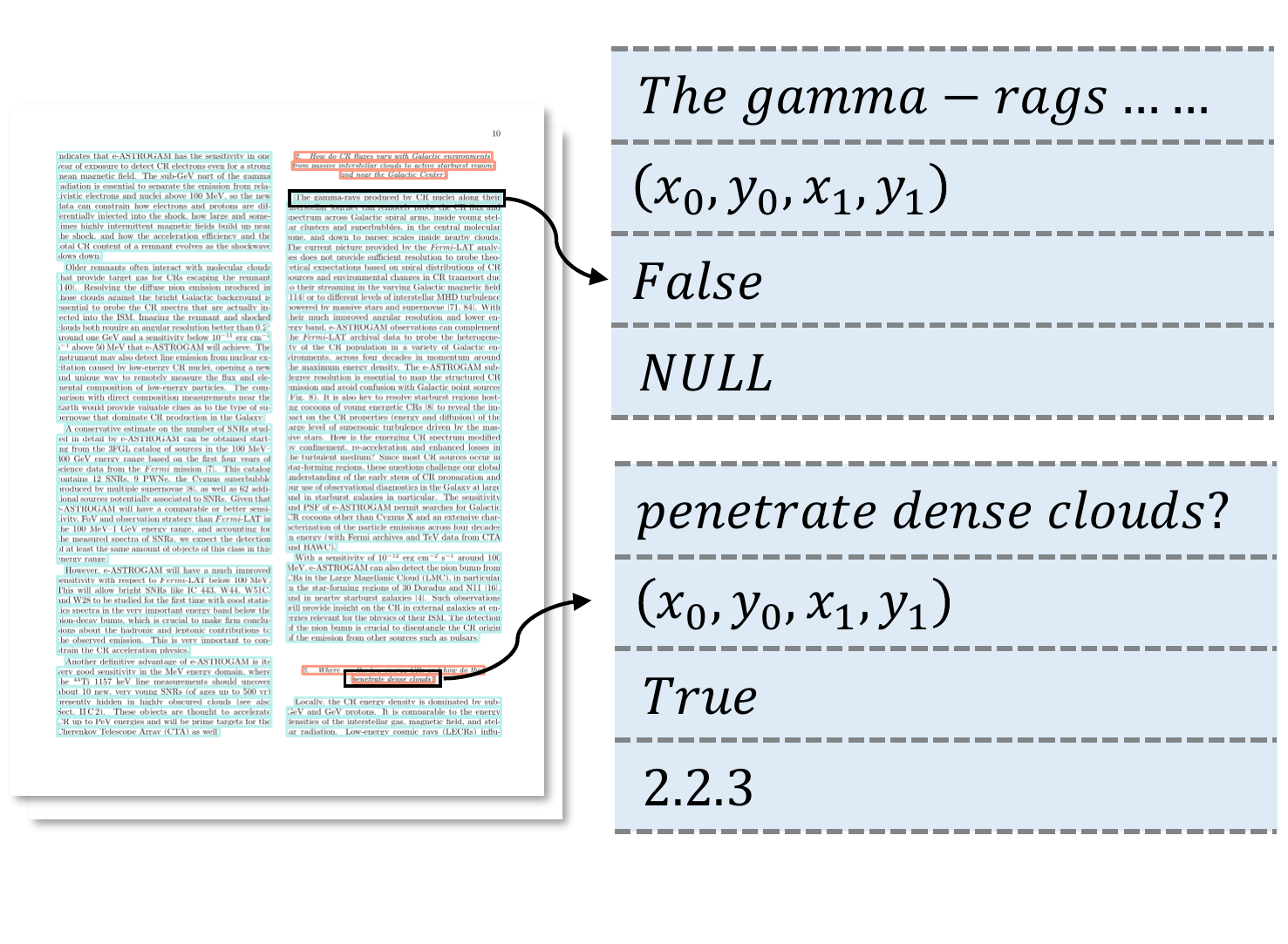}}		
	  \\
	\caption{\textit{HierDoc} contains two kinds of annotations. Document-level annotations serve as targets of ToC extraction. Entity-level annotations provide a quadruple for each entity, $(content, position, heading, id)$. $heading$ is a boolean variable indicating whether an entity is a heading and $id$ identifies the hierarchical position of the entity in the ToC tree.}
    \label{figure:dataset}
\end{figure}

The document-level annotation, as shown in Fig. \ref{figure:dataset}, can be parsed using regular expressions with LaTeX source code. It is used as the target during the test phase. We also generate a quadruple $(content, position, heading, id)$ for each entity. We parse documents in PDF format using pdfplumber\footnote{https://github.com/jsvine/pdfplumber} to obtain the text contents with positions for each entity. After that, the entities are matched with document-level annotations to obtain $heading$ and $id$ according to the Word Error Rate (WER). To facilitate model processing, the entities on each page are organized from top-to-bottom and left-to-right. Overall, $HierDoc$ provides document images, entity-level annotations, and document-level annotations.

\section{THE PROPOSED APPROACH}
\begin{figure*}
\centerline{\includegraphics[width=1\linewidth]{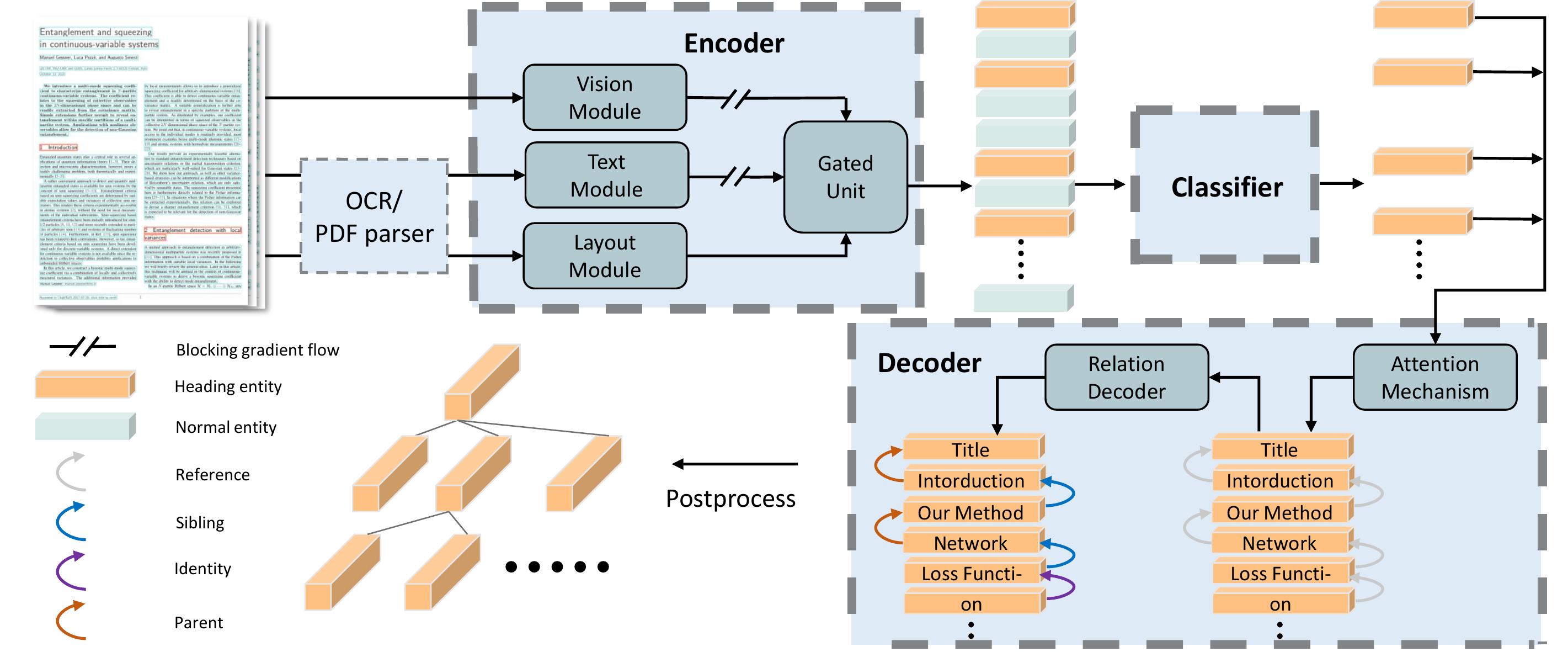}}
\caption{Architecture of the MTD for ToC extraction.}
\label{figure:model}
\end{figure*}

The overall pipeline of MTD is shown in Fig. \ref{figure:model}. MTD consists of three components: encoder, classifier and decoder. The vision module, text module, and layout module are first applied to the input document to extract features of each entity. Then the gated unit in the encoder is used to obtain the multimodality features. The following classifier detects the heading entities and feeds them to the decoder. Finally, the decoder parses the hierarchy of headings to produce the ToC. In the following subsections, we will first formalize our method and then elaborate on the components of MTD.

\subsection{Formalization\label{sub:formalization}}
Given a document, we first obtain the layout and plain text of entities by OCR engines or PDF Parsers, as discussed in Section \ref{sec:dataset}. Then, the encoder extracts the features of all entities. Next, the classifier divides them into two categories, \textit{heading entities} or \textit{normal entities}. A heading entity is one of the headings of the document, while a normal entity is not. Finally, starting from an empty tree with a single root node, all heading entities are iterated in the sequence of how they appear within the document to generate the ToC tree.

More specifically, there may be one of three relationships between two heading entities, namely  \textit{parent}, \textit{sibling} and \textit{identity}. As shown in Fig. \ref{figure:formal}, \textit{Introduction}, \textit{Method} and \textit{Conclusion} are chapters in \textit{TITLE}. Intuitively, the relationship between \textit{Introduction} and \textit{TITLE} is \textit{parent}. For chapter \textit{Method}, we predict the relationship \textit{sibling} between it and \textit{Introduction}. It can avoid predicting the relationship between two entities that are far apart in the document, since the \textit{TITLE} may have many children. We design the relationship \textit{identity} because the heading may be divided into several entities due to the limited length of text lines. So far, there is one critical issue that remains addressed. For the current entity, which entity should be used to predict the relationship with it? The answer is the \textit{reference entity}. The reference entity is the entity immediately preceding the current entity within the document, between which one of the relationships described above exists. Finally, after obtaining the reference entity and relationship for each heading entity, we can generate the ToC tree simply.

\subsection{Encoder}
Different from previous works\cite{2018content, 2007content, 2015hybrid}, the encoder requires no hand-crafted features to extract features of each entity. Firstly, with the entity-level annotations, the encoder extracts the visual features $f^{v}_{t}$, the textual features $f^{s}_{t}$ and position features $f^{p}_{t}$ of each entity. Then, before being fed to the next module, $f^v_t$, $f^s_t$, and $f^p_t$ are fused into $f_t$. $f_{t} \in \mathbb{R}^{d}$, $f^{v}_{t} \in \mathbb{R}^{d}$, $f^{s}_{t} \in \mathbb{R}^{d}$, $f^{p}_{t} \in \mathbb{R}^{8}$, $t \in [1, N]$, and $N$ is the number of entities in the document.

\subsubsection{Vision Module} The vision module takes document images as input. It uses FPN\cite{fpn} to aggregate feature maps from ResNet-34\cite{resnet}, then pools a fixed-size feature map $\hat{f}^{v}_{t}$ with the RoIAlign\cite{roialign} for each entity. In order to integrate with features from other domains, the 2-dimensional feature map $\hat{f}^{v}_{t}$ is flatten to produce visual features $f^{v}_{t}$. The ResNet-34 and FPN are pretrained on 1000 scientific papers with a text detection task\cite{FCOS}.

\subsubsection{Text Module} Recent studies\cite{layoutlm} \cite{layoutreader} show that textual features play an extremely important role in documents. Following \cite{sem}, the BERT is used to extract the textual features $f^{s}_{t}$. To make the extracted semantic features more suitable for our network, two linear transformations with a RELU activation are added following the BERT. It is worth noting that both BERT and ResNet-34+FPN do not update their parameters during the training phase to save GPU memory and improve training speed.

\subsubsection{Layout Module} The layout module generates $f^{p}_{t}$  as the following:
$$ {f}^{p}_{t}=(\frac{x_t^{lt}}{{W}}, \frac{y_t^{lt}}{{H}}, \frac{x_t^{rb}}{{W}}, \frac{y_t^{rb}}{{H}}, \frac{w_t}{\bar{w}}, \frac{h_t}{\bar{ h}}, \frac{y_t^{lt} - y_{t-1}^{rb}}{\bar{h}}, \frac{y_{t+1}^{lt} - y_{t}^{rb}}{\bar{h}})$$
$w_t$, $h_t$, $(x_t^{lt} , y_t^{lt}, x_t^{rb} , y_t^{rb})$ correspond to the width, the height, and the coordinate position of the bounding box of each entity. ${W}$, ${H}$ denote the width and height of the whole page. We normalize $h_t$ with $\bar{h}$, which is the average height of all bounding boxes. We believe that $\bar{h}$ is a better measure to normalize $h$ than the commonly used $H$, because it reflects the difference in spatial height between entities better.

\subsubsection{Gated Unit\label{sec:fuse}}
Inspired by \cite{gateunit} \cite{srn}, we introduce some trainable weights to balance the contribution of $ {f}^{v}_t$ , $ {f}^{s}_t$, and $ {f}^{p}_t$. The fused feature $f_t$ is produced as follows:

$$\left\{\begin{aligned}
z_{t} &=\sigma\left(\mathbf{W}_{z} \cdot\left[f^{v}_t, f^{s}_t,f^{p}_t\right]\right) \\
f_{t} &=z_{t} * f^{v}_t+\left(1-z_t\right) * f^{s}_t + \mathbf{E}_{z}f^p_t
\end{aligned}\right.$$
Where $\mathbf{W}_{z}$, $\mathbf{E}_z$ are trainable weight matrices, and $\sigma$ is the sigmoid function, $\mathbf{W}_{z} \in \mathbb{R}^{d \times (2d+8)}$, $\mathbf{E}_{z} \in \mathbb{R}^{d \times 8}$. The fused features $f_t$ are the final output of the encoder.

\subsection{Classifier\label{sub:class}}
The classifier divides all entities into two categories, i.e., heading entities and normal entities, as discussed in Section \ref{sub:formalization}.

Before classification, Bidirectional Gated Recurrent Unit (BiGRU)\cite{gru} is used to capture global information:
$$g_t= {\rm{BiGRU}}(f_t, g_{t-1}, g_{t+1})$$
where $g_t$ is the hidden state. Then we apply a fully connected layer and a softmax activation to classify each entity:
$$x_t={\rm softmax}(\mathbf{W}_{c}g_t+\mathbf{b}_{c})$$
where $x_t$ represents the classification score, $x_t \in \mathbb{R}^{2}$, $\mathbf{W}_{c} \in \mathbb{R}^{2 \times d}$, $\mathbf{b}_{c} \in \mathbb{R}^{2}$. Considering there is an imbalance between heading entities and normal entities, we define the classification loss as follows: 
$$L_{cls}=\frac{1}{N}\sum_{t \le N} {\rm FL}(x_t, p_t)$$
where $p_t$ is the class label of the entity and {\rm FL} is the focal loss proposed in \cite{focalloss} to deal with the problem of data imbalance existing between various classes.

\subsection{Decoder}
The decoder takes detected heading entities as input and predicts the relationship between them to parse the heading entities into a tree structure. The features of heading entities can be denoted as $\hat{M}$, where $\hat{M} \in \mathbb{R}^{C \times d}$, and $C$ is the number of the heading entities. So far, the features of each heading entity are still independent of each other. Therefore, we introduce the transformer\cite{transformer} to capture long-range dependencies on heading entities. We take the features $\hat{M}$ as query, key and value, which are required by the transformer. The output of the transformer as the final features ${M}$ has a global receptive field.

Inspired by the successful applications of attention mechanism\cite{wap} \cite{sem}, we build the attention mechanism into the decoder. To find the reference entity, we compute the prediction of the current hidden state $\hat{h_s}$ from previous context vector $c_{s-1}$ and its hidden state $h_{s-1}$ with a GRU:

$$\hat h_s={\rm{GRU}}(c_{s-1}, h_{s-1})$$

Then we employ an attention mechanism with $\hat{h}_s$ as the query and the heading entity features $M$ as both key and value:

$$e_s = {\rm Attn}(M, \hat{h}_s)$$
$$c_s = \frac{e_s}{\Vert e_s \Vert_1}M$$
where $\Vert \cdot \Vert_1$ is the vector 1-norm. We define ${\rm Attn}$ function as follows:

$${D}={Q} * \sum_{l=1}^{s-1} {e}_{l}$$
$$\hat{e}_{si} = v^T{\rm tanh}(W_{h}\hat{h}_s + W_mm_i + W_dd_i)$$
$$e_{si} = {\rm activate}(e_{si}, e_s)$$
in which
$$\operatorname{activate}(e_{si}, e_s)=\left\{\begin{array}{ll}
1 & \text { if } i=\mathop{\arg\max}\limits_{j} (e_{sj})\\
0 & \text { otherwise }
\end{array}\right.$$
where $*$ denotes a convolution layer, $\sum_{l=1}^{s-1} {e}_{l}$ denotes the sum of the past determined reference entity, $ \hat{e}_{s,i} $ denotes the output energy, $d_{i}$ denotes the element of $D$, which is used to keep track of past alignment information. It is worth noting that the attention mechanism is completed on the features of each heading entity. Considering that there is only one reference entity for each heading entity, we use $\operatorname{activate}$ to obtain the attention probability instead of softmax, which means that the weight of the reference entity is set to 1 while other entities remain 0.

With the context vector $c_s$, we compute the current hidden state:
$$h_s = \operatorname{GRU}(c_s, \hat{h}_s)$$

The training loss of locating the reference entity is defined as:

$$L_{ref} = \frac{1}{M}\sum_{s \le M}\operatorname{FL}(\hat{e_s}, y_s)$$
where loss function $\operatorname{FL}$ has been defined in Section \ref{sub:class} and $y_s$ denotes the ground truth of the reference entity for time step $s$. $y_{si}$ is 1 if $i^{th}$ heading entity is the reference entity of the current heading entity, otherwise 0.

The relationship between the current entity and its reference entity is predicted through an FFN\cite{transformer}:

$$r_s = {{\rm FFN}}([c_s, h_s])$$

$$\operatorname{FFN}({x})=\mathbf{W}_{2}\max \left(0,  \mathbf{W}_1{x}+\mathbf{b}_{1}\right) + \mathbf{b}_{2}$$
where FFN is actually two linear transformations with a ReLU activation
in between. $r_s \in \mathbb{R}^{3}$, $\mathbf{W}_{1} \in \mathbb{R}^{d_{in} \times 2d}$, $\mathbf{b}_{1} \in \mathbb{R}^{d_{in}}$, $\mathbf{W}_{2} \in \mathbb{R}^{3 \times d_{in}}$, $\mathbf{b}_{2} \in \mathbb{R}^{3}$.

The loss of predicting relationships is as follows:

$$L_{re} = \frac{1}{M}\sum_{s \le M}\operatorname{FL}({r_s}, q_s)$$
where $q_s$ denotes the ground truth of the relation between the current heading entity and its reference entity at time step $s$.

\section{EXPERIMENTS}

\subsection{Metric}
In this paper, we use both the F1-Measure and Tree-Edit-Distance-based Similarity (TEDS) metric\cite{teds} to evaluate the performance of our model for ToC extraction. 

To use the F1-Measure, the relationships among the heading entities and corresponding reference entities need to be extracted. Then F1-Measure measures the percentage of correctly extracted pairs of heading entities, where heading entities and the relationships between them are the same as the ground truth.

While using the TEDS metric, we need to present ToC as a tree structure. The tree-edit distance\cite{ted} is used to measure the similarity between two trees. TEDS is computed as:

$$\operatorname{TEDS}\left(T_{a}, T_{b}\right)=1-\frac{\operatorname{EditDist}\left(T_{a}, T_{b}\right)}{\max \left(\left|T_{a}\right|,\left|T_{b}\right|\right)}$$
where $T_a$ and $T_b$ are ToC trees. $\operatorname{EditDist}$ represents the tree-edit distance, and $\left|T\right|$ is the number of nodes in $T$.

\subsection{Implementation Details\label{sub:detail}}
In the vision module, ResNet-34+FPN is pre-trained on 1000 scientific documents for a text detection task \cite{FCOS}. The pool size of RoIAlign is set to 3 $\times$ 3. The BERT used in the text module is available on GitHub\footnote{ https://github.com/huggingface/transformers}. The channel number of visual, textual, and fused features is 128. The number of stacks for the transformer in the decoder is set to 3. And the hidden state dimension in the classifier and decoder is both 128, too.

The training objective of our model is to minimize the function:
$$O = \alpha_1L_{cls} + \alpha_2L_{ref} +  \alpha_3L_{re}$$
where $\alpha_1$, $\alpha_2$ and $\alpha_3$ are set to 1. Random scale is adopted during training. Both BERT and ResNet-34+FPN do not update their parameters to save GPU memory. We employ the Adam algorithm \cite{adam} for optimization, with the following hyper parameters: $\beta_1$= 0.9 and $\beta_2$ = 0.999. We set the learning rate using the cosine annealing schedule \cite{cosineLR}. The minimum learning rate and the initial learning rate are set to 1e-6 and 5e-4, respectively. The batch size varies between 1 and 4 according to the number of pages of the document. All experiments are implemented with a single Tesla V100 GPU with 32GB RAM Memory. The whole framework is implemented using PyTorch.

\subsection{Ablation Study}
\subsubsection{The Effectiveness of Each Modality}
We perform ablation studies to evaluate the effectiveness of each modality in this section. The \textit{Heading Detecting} in tables represents the task of detecting heading entities, and the metric we use is F1-Measure. As shown in Table \ref{tablemodal}, 
when any of the visual, textual, or layout modalities is removed, the performance of the model drops. This suggests that multimodal features play an important role in ToC extraction. Note that the parameters of our visual and textual modules are frozen during training. Therefore, they need to be properly pre-trained as discussed in Section \ref{sub:detail}.


\begin{table}[]
\caption{The Effectiveness of Each Modality.}
\label{tablemodal}
\centering
\begin{tabular}{c|ccc|cc}
\hline
\multirow{2}{*}{Method} & \multicolumn{3}{c|}{Heading Detecting} & \multicolumn{2}{c}{ToC Extraction} \\ \cline{2-6} 
 & \multicolumn{1}{c|}{P} & \multicolumn{1}{c|}{R} & F1 & TEDS & F1 \\ \hline
w/o Text & \multicolumn{1}{c|}{88.6} & \multicolumn{1}{c|}{83.0} & 85.7 & 63.7 & 62.5 \\
w/o Layout & \multicolumn{1}{c|}{95.0} & \multicolumn{1}{c|}{89.2} & 92.0 & 80.4 & 79.6 \\
w/o Vision & \multicolumn{1}{c|}{\textbf{97.5}} & \multicolumn{1}{c|}{93.7} & 95.5 & 86.5 & 87.7 \\ \hline
MTD & \multicolumn{1}{c|}{96.2} & \multicolumn{1}{c|}{\textbf{96.0}} & \textbf{96.1} & \textbf{87.2} & \textbf{88.1} \\ \hline
\end{tabular}
\end{table}

\subsubsection{The Effectiveness of Feature Fusion Strategies}

We introduce a novel feature fusion strategy in Section \ref{sec:fuse}. In this section, we conduct experiments to compare it with three common feature fusion strategies, including dot, concatenate, and add. Table \ref{tabel:fuse} shows that the Gated Unit outperforms other strategies. Thus, the Gated Unit is utilized in our approach.


\begin{table}[]
\caption{The Effectiveness of Feature Fusion Strategies.}
\label{tabel:fuse}
\centering
\begin{tabular}{c|ccc|cl}
\hline
\multirow{2}{*}{Method} & \multicolumn{3}{c|}{Heading Detecting} & \multicolumn{2}{c}{ToC Extraction} \\ \cline{2-6} 
 & \multicolumn{1}{c|}{P} & \multicolumn{1}{c|}{R} & F1 & \multicolumn{1}{c|}{TEDS} & \multicolumn{1}{c}{F1} \\ \hline
Dot & \multicolumn{1}{c|}{96.1} & \multicolumn{1}{c|}{93.9} & 95.0 & \multicolumn{1}{c|}{84.6} & 84.9 \\
Concat & \multicolumn{1}{c|}{96.8} & \multicolumn{1}{c|}{93.8} & 95.3 & \multicolumn{1}{c|}{84.6} & 85.3 \\
Add & \multicolumn{1}{c|}{\textbf{97.0}} & \multicolumn{1}{c|}{94.0} & 95.5 & \multicolumn{1}{c|}{85.7} & 86.3 \\ \hline
Gated Unit & \multicolumn{1}{c|}{96.2} & \multicolumn{1}{c|}{\textbf{96.0}} & \textbf{96.1} & \multicolumn{1}{c|}{\textbf{87.2}} & \textbf{88.1} \\ \hline
\end{tabular}
\end{table}

\subsubsection{The Effectiveness of Decoding Methods.}
To verify the performance of the decoder in MTD, we replace it with a \textit{C}-class classifier, which predicts the depth of headings. \textit{C} is set to 5 in \textit{HierDoc}. The transformer unit is reserved for a fair comparison.

\begin{table}[h]
\caption{The Effectiveness of Decoding Methods.}
\label{tabel:decoder}
\centering
\begin{tabular}{c|c}
\hline
Method & ToC Extraction (TEDS) \\ \hline
C-class Classifier & 72.1 \\ \hline
Decoder & \textbf{87.2} \\ \hline
\end{tabular}
\end{table}

As described in Table \ref{tabel:decoder}, the \textit{C}-class Classifier leads to degradation of performance to a certain extent. Intuitively, the Decoder in MTD is more suitable for parsing the hierarchy of entities into a tree structure. In addition, the Decoder can generate a tree of any depth by defining three types of relationships, namely \textit{parent}, \textit{sibling}, and \textit{identity}, while the \textit{C}-class classifier has to restrict the tree to a depth of \textit{C}.

\section{Conclusion}
In this paper, we release a dataset \textit{HierDoc} for deep learning based methods and further propose the MTD for ToC extraction. \textit{HierDoc} contains 650 document images and annotations in both the document-level and entity-level. The document-level annotations serve as the target of ToC extraction in the test phase, while the entity-level annotations are generated for training. MTD mainly consists of three parts, i.e., encoder, classifier, and decoder. The encoder extracts features of entities, then the classifier selects the heading entities and the decoder parses the hierarchy of heading entities. We demonstrate that the use of multimodality features of vision, text, and layout in the encoder boosts model performance. The gated unit used in the encoder also outperforms other feature fusion strategies. The decoder builds the ToC tree by parsing the hierarchy of heading entities. The decoder outperforms the vanilla \textit{C}-class classifier by a large margin, and it can generate a tree of any depth. MTD achieves an average TEDS of 87.2\%  and an average F1-Measure of 88.1\% on the test set of HierDoc. We hope that our MTD could serve as a strong baseline for deep learning based ToC extraction in the future.



\bibliographystyle{IEEETran}
\bibliography{refs_icpr}

\begin{thebibliography}{10}
\providecommand{\url}[1]{#1}
\csname url@samestyle\endcsname
\providecommand{\newblock}{\relax}
\providecommand{\bibinfo}[2]{#2}
\providecommand{\BIBentrySTDinterwordspacing}{\spaceskip=0pt\relax}
\providecommand{\BIBentryALTinterwordstretchfactor}{4}
\providecommand{\BIBentryALTinterwordspacing}{\spaceskip=\fontdimen2\font plus
\BIBentryALTinterwordstretchfactor\fontdimen3\font minus
  \fontdimen4\font\relax}
\providecommand{\BIBforeignlanguage}[2]{{%
\expandafter\ifx\csname l@#1\endcsname\relax
\typeout{** WARNING: IEEEtran.bst: No hyphenation pattern has been}%
\typeout{** loaded for the language `#1'. Using the pattern for}%
\typeout{** the default language instead.}%
\else
\language=\csname l@#1\endcsname
\fi
#2}}
\providecommand{\BIBdecl}{\relax}
\BIBdecl

\bibitem{2020Tocpages}
M.~El-Haj, P.~Alves, P.~Rayson, M.~Walker, and S.~Young, ``Retrieving,
  classifying and analysing narrative commentary in unstructured (glossy)
  annual reports published as pdf files,'' \emph{Accounting and Business
  Research}, vol.~50, no.~1, pp. 6--34, 2020.

\bibitem{2017Tocpages}
A.~Doucet, M.~Coustaty \emph{et~al.}, ``Enhancing table of contents extraction
  by system aggregation,'' in \emph{2017 14th IAPR international conference on
  document analysis and recognition (ICDAR)}, vol.~1.\hskip 1em plus 0.5em
  minus 0.4em\relax IEEE, 2017, pp. 242--247.

\bibitem{noToC}
A.~Doucet, G.~Kazai, B.~Dresevic, A.~Uzelac, B.~Radakovic, and N.~Todic,
  ``Setting up a competition framework for the evaluation of structure
  extraction from ocr-ed books,'' \emph{International Journal on Document
  Analysis \& Recognition}, vol.~14, no.~1, pp. 45--52, 2011.

\bibitem{2018content}
A.~A.~M. Gopinath, S.~Wilson, and N.~Sadeh, ``Supervised and unsupervised
  methods for robust separation of section titles and prose text in web
  documents,'' in \emph{Proceedings of the 2018 Conference on Empirical Methods
  in Natural Language Processing}, 2018, pp. 850--855.

\bibitem{2007content}
A.~M. Namboodiri and A.~K. Jain, ``Document structure and layout analysis,'' in
  \emph{Digital Document Processing}.\hskip 1em plus 0.5em minus 0.4em\relax
  Springer, 2007, pp. 29--48.

\bibitem{2015hybrid}
S.~Tuarob, P.~Mitra, and C.~L. Giles, ``A hybrid approach to discover semantic
  hierarchical sections in scholarly documents,'' in \emph{2015 13th
  international conference on document analysis and recognition (ICDAR)}.\hskip
  1em plus 0.5em minus 0.4em\relax IEEE, 2015, pp. 1081--1085.

\bibitem{layoutlm}
Y.~Xu, M.~Li, L.~Cui, S.~Huang, F.~Wei, and M.~Zhou, ``Layoutlm: Pre-training
  of text and layout for document image understanding,'' in \emph{Proceedings
  of the 26th ACM SIGKDD International Conference on Knowledge Discovery \&
  Data Mining}, 2020, pp. 1192--1200.

\bibitem{devlin2018bert}
J.~Devlin, M.-W. Chang, K.~Lee, and K.~Toutanova, ``Bert: Pre-training of deep
  bidirectional transformers for language understanding,'' \emph{arXiv preprint
  arXiv:1810.04805}, 2018.

\bibitem{vibertgrid}
W.~Lin, Q.~Gao, L.~Sun, Z.~Zhong, K.~Hu, Q.~Ren, and Q.~Huo, ``Vibertgrid: A
  jointly trained multi-modal 2d document representation for key information
  extraction from documents,'' \emph{arXiv preprint arXiv:2105.11672}, 2021.

\bibitem{isri}
T.~A. Nartker, S.~V. Rice, and S.~E. Lumos, ``Software tools and test data for
  research and testing of page-reading ocr systems,'' in \emph{Document
  Recognition and Retrieval XII, 16-20 January 2005, San Jose, California, USA,
  Proceedings}, 2005.

\bibitem{marg}
G.~Thoma, ``The national library of medicine,'' in
  \emph{http://marg.nlm.nih.gov/, Bethesda, USA}, 2005.

\bibitem{2015icdar}
C.~Clausner, C.~Papadopoulos, S.~Pletschacher, and A.~Antonacopoulos, ``The enp
  image and ground truth dataset of historical newspapers,'' in \emph{2015 13th
  International Conference on Document Analysis and Recognition (ICDAR)}, 2015,
  pp. 931--935.

\bibitem{2017icdar}
C.~Clausner, A.~Antonacopoulos, and S.~Pletschacher, ``Icdar2017 competition on
  recognition of documents with complex layouts - rdcl2017,'' in \emph{2017
  14th IAPR International Conference on Document Analysis and Recognition
  (ICDAR)}, vol.~01, 2017, pp. 1404--1410.

\bibitem{2009icdar}
A.~Antonacopoulos, D.~Bridson, C.~Papadopoulos, and S.~Pletschacher, ``A
  realistic dataset for performance evaluation of document layout analysis,''
  in \emph{2009 10th International Conference on Document Analysis and
  Recognition}.\hskip 1em plus 0.5em minus 0.4em\relax IEEE, 2009, pp.
  296--300.

\bibitem{publaynet}
X.~Zhong, J.~Tang, and A.~Jimeno~Yepes, ``Publaynet: Largest dataset ever for
  document layout analysis,'' in \emph{2019 International Conference on
  Document Analysis and Recognition (ICDAR)}, 2019, pp. 1015--1022.

\bibitem{data:fin}
N.-I. Bentabet, R.~Juge, I.~El~Maarouf, V.~Mouilleron,
  D.~Valsamou-Stanislawski, and M.~El-Haj, ``The financial document structure
  extraction shared task (fintoc 2020),'' in \emph{Proceedings of the 1st Joint
  Workshop on Financial Narrative Processing and MultiLing Financial
  Summarisation}, 2020, pp. 13--22.

\bibitem{teds}
X.~Zhong, E.~ShafieiBavani, and A.~Jimeno~Yepes, ``Image-based table
  recognition: data, model, and evaluation,'' in \emph{Computer Vision--ECCV
  2020: 16th European Conference, Glasgow, UK, August 23--28, 2020,
  Proceedings, Part XXI 16}.\hskip 1em plus 0.5em minus 0.4em\relax Springer,
  2020, pp. 564--580.

\bibitem{fpn}
T.-Y. Lin, P.~Doll{\'a}r, R.~Girshick, K.~He, B.~Hariharan, and S.~Belongie,
  ``Feature pyramid networks for object detection,'' in \emph{Proceedings of
  the IEEE conference on computer vision and pattern recognition}, 2017, pp.
  2117--2125.

\bibitem{resnet}
K.~He, X.~Zhang, S.~Ren, and J.~Sun, ``Deep residual learning for image
  recognition,'' in \emph{Proceedings of the IEEE conference on computer vision
  and pattern recognition}, 2016, pp. 770--778.

\bibitem{roialign}
K.~He, G.~Gkioxari, P.~Doll{\'a}r, and R.~Girshick, ``Mask r-cnn,'' in
  \emph{Proceedings of the IEEE international conference on computer vision},
  2017, pp. 2961--2969.

\bibitem{FCOS}
Z.~Tian, C.~Shen, H.~Chen, and T.~He, ``Fcos: Fully convolutional one-stage
  object detection,'' in \emph{2019 IEEE/CVF International Conference on
  Computer Vision (ICCV)}, 2020.

\bibitem{layoutreader}
Z.~Wang, Y.~Xu, L.~Cui, J.~Shang, and F.~Wei, ``Layoutreader: Pre-training of
  text and layout for reading order detection,'' \emph{arXiv preprint
  arXiv:2108.11591}, 2021.

\bibitem{sem}
Z.~Zhang, J.~Zhang, and J.~Du, ``Split, embed and merge: An accurate table
  structure recognizer,'' \emph{arXiv preprint arXiv:2107.05214}, 2021.

\bibitem{gateunit}
J.~Arevalo, T.~Solorio, M.~Montes-y G{\'o}mez, and F.~A. Gonz{\'a}lez, ``Gated
  multimodal units for information fusion,'' \emph{arXiv preprint
  arXiv:1702.01992}, 2017.

\bibitem{srn}
D.~Yu, X.~Li, C.~Zhang, T.~Liu, J.~Han, J.~Liu, and E.~Ding, ``Towards accurate
  scene text recognition with semantic reasoning networks,'' in
  \emph{Proceedings of the IEEE/CVF Conference on Computer Vision and Pattern
  Recognition}, 2020, pp. 12\,113--12\,122.

\bibitem{gru}
J.~Chung, C.~Gulcehre, K.~Cho, and Y.~Bengio, ``Empirical evaluation of gated
  recurrent neural networks on sequence modeling,'' \emph{arXiv preprint
  arXiv:1412.3555}, 2014.

\bibitem{focalloss}
T.~Y. Lin, P.~Goyal, R.~Girshick, K.~He, and P.~Dollar, ``Focal loss for dense
  object detection,'' in \emph{2017 IEEE International Conference on Computer
  Vision (ICCV)}, 2017.

\bibitem{transformer}
A.~Vaswani, N.~Shazeer, N.~Parmar, J.~Uszkoreit, L.~Jones, A.~N. Gomez,
  {\L}.~Kaiser, and I.~Polosukhin, ``Attention is all you need,'' in
  \emph{Advances in neural information processing systems}, 2017, pp.
  5998--6008.

\bibitem{wap}
J.~Zhang, J.~Du, S.~Zhang, D.~Liu, Y.~Hu, J.~Hu, S.~Wei, and L.~Dai, ``Watch,
  attend and parse: An end-to-end neural network based approach to handwritten
  mathematical expression recognition,'' \emph{Pattern Recognition}, vol.~71,
  pp. 196--206, 2017.

\bibitem{ted}
M.~Pawlik and N.~Augsten, ``Tree edit distance: Robust and memory-efficient,''
  \emph{Information Systems}, vol.~56, pp. 157--173, 2016.

\bibitem{adam}
D.~Kingma and J.~Ba, ``Adam: A method for stochastic optimization,''
  \emph{Computer Science}, 2014.

\bibitem{cosineLR}
I.~Loshchilov and F.~Hutter, ``Sgdr: Stochastic gradient descent with warm
  restarts,'' \emph{arXiv preprint arXiv:1608.03983}, 2016.

\end{thebibliography}

\end{document}